# Principles of Explanation in Human-AI Systems

Shane T. Mueller[1], Elizabeth S. Veinott[2], Robert R. Hoffman[3], Gary Klein[4], Lamia Alam[5], Tauseef Mamun[6], and William J. Clancey[7]

Michigan Technological University,[1,2,5,6] Institute for Human and Machine Cognition,[3,7] Macrocognition, LLC[4]
shanem@mtu.edu,[1] eveinott@mtu.edu,[2] rhoffman@ihmc.us,[3] gary@macrocognition.com,[4] lalam@mtu.edu,[5] tmamun@mtu.edu,[6] wclancey@ihmc.us[7]

**Abstract**

*Explainable Artificial Intelligence (XAI) has re-emerged in response to the development of modern AI and ML systems. These systems are complex and sometimes biased, but they nevertheless make decisions that impact our lives. XAI systems are frequently algorithm-focused; starting and ending with an algorithm that implements a basic untested idea about explainability. These systems are often not tested to determine whether the algorithm helps users accomplish any goals, and so their explainability remains unproven. We propose an alternative: to start with human-focused principles for the design, testing, and implementation of XAI systems, and implement algorithms to serve that purpose. In this paper, we review some of the basic concepts that have been used for user-centered XAI systems over the past 40 years of research. Based on these, we describe the "Self-Explanation Scorecard", which can help developers understand how they can empower users to by enabling self-explanation. Finally, we present a set of empirically-grounded, user-centered design principles that may guide developers to create successful explainable systems.*

## User-Centered Explanation in AI

Although usability testing is a cornerstone of user-centered design, evaluation often comes too late to provide guidance about implementing a usable system. In response, researchers and designers have proposed guidelines that codify research on human users and advocate for the involvement of users in system development from the beginning (e.g., Greenbaum and Kyng 1991; Hoffman et al. 2010). The most famous and detailed set of guidelines may be Apple's Human Interface Guidelines (cf. Mountford 1998), but others have proposed simpler principles such as Neilson's (1994) interface design heuristics or Karat's (1998) "User's Bill of Rights".

With the advent of new, powerful AI systems that are complex and difficult to understand, the field of Explainable AI (XAI) has re-emerged as an important area of human-machine interaction. Much of the interest in XAI has focused on deep learning systems. Consequently, most explanations have concentrated on technologies to visualize or otherwise expose deep networks structures, features, or decisions, or the large data sets they are trained on. Currently, new "explainable" systems often begin with an assertion about what makes for a good explanation (for example, "an explanation should be simple"), and end with a computational architecture implementing that idea (mapping a complex network onto a simpler structure that serves as the explanation), without ever providing evidence either that the idea was justified, or that the resulting system is effective. This work has often ignored the insights from more than 40 years of human-centered research on explanation in AI systems, which began in the early days of expert systems (see Swartout 1977; Clancey 1986; Moore and Swartout 1988) and continued in the work on intelligent tutoring systems (Woolf 2007). Some researchers within the XAI community have advocated for a principled and rigorous basis for explanatory systems based on social and cognitive sciences (e.g., Miller 2017; Doshi-Velez and Kim 2017; Ribera and Lapedriza 2019). Furthermore, our research team, with a background in human factors psychology and cognitive systems engineering, has proposed systematic approaches to evaluating the effectiveness of explanations (cf. Hoffman et al. 2017; 2018abc). But lessons from these psychological perspectives appear to remain overshadowed by algorithm-centered approaches to XAI. Thus, the explanation capabilities needed for genuine human-AI collaboration are underdeveloped, and often non-existent. This is a gap that cannot be addressed without systematic use-inspired research.

Thus, there is a need for use-inspired human-focused guidelines for XAI. Indeed, there are now decades of scholarship about how to best implement human-centered systems (Hoffman 2012) including AI (Clancey 2020). Yet many AI developers understandably are not well-versed in the methods of human-centered design. Thus, the purpose of this review is to lay out the major themes relevant to human-centered development of explainable AI systems. On the basis of this review, we will then describe the Self-Explanation Scorecard; a way of understanding how an AI system might support self-explanation of its users. We conclude with a list of human-centered principles for designing explainable AI.

Table 1. Concepts related to the development of explainable AI systems.

| Concept | Reference | Relevance |
|---|---|---|
| **Measurement Domains** | | |
| Explanation Goodness | Hoffman et al. 2018a | Simple, *a priori* assessment for what makes for a good explanation, typically based on philosophical literature. |
| Explanation Satisfaction | Hoffman et al. 2018a | Measured user satisfaction with an explanation. |
| Justified or Appropriate Trust | Hoffman 2017 | Explanation promotes justified trust and mistrust and mitigates unjustified trust and mistrust. |
| Mental Models | Kulesza et al. 2015 | Explanation improves user's mental model and understanding. |
| Performance | Hoffman et al. 2018a | Explanation and enhances performance of joint human-AI systems. |
| Licensure | Tate et al. 2016 | A licensure framework permits appropriate future use of AI. |
| **Features/Properties of an Explainable Architecture** | | |
| User Models | Kass and Finin 1988 | Consider and incorporate predictive models of the user. |
| User Context | Brézillon 1994; Doyle et al. 2003 | User context includes goals, novice/expert status, environment, etc. |
| Relevant | Sørmo et al. 2005; Lim & Dey 2009 | Explanation is needed in response to relevant situations, anomalies, and violations of expectation. |
| Timely | Alam 2020 | Explanations may only help at critical times (errors, status changes). |
| Experiential | Mueller and Klein 2011 | Experiential training is necessary for promoting understanding of intelligent machine systems. |
| Iterative, Interactive, and Explorable | Lakkaraju et al. 2017; Hoffman et al. 2020 | Explanations should be explorable by users. |
| **Properties/Formats of Explanations** | | |
| Explanation Focus (Global versus Local) | Wick and Thompson 1992 | Explanations can focus on how a system works (global) or why it made a particular decision (local/justification). |
| Logic/rationale | Swartout 1977; Hendricks et al. 2016 | Explanations can involve verbal and logical description of reasoning for a decision or process. |
| Examples and Cases | Doyle et al. 2003; Yang and Shafto 2017 | Examples can be useful, but must be selected appropriately. |
| Counterfactual and Contrastive Reasoning | Miller 2017 | An explanation typically involves a contrast, comparison, or counterfactual reasoning. |
| Feature importance or outcome likelihood | Ribeiro et al. 2016 | Highlighting features important for a decision help justify specific decisions and create a user's mental model of the system. |
| Boundary conditions | Mueller and Klein 2011 | Understanding is improved when humans understand competence envelope of the system (i.e.,, its boundary conditions.) |
| Self or other-explanation and Critique | Chi, et al. 1989; Veinott, et. al. 2010 | Critique and self or other explanation are effective ways to boost understanding of complex processes and intelligent systems. |
| Persuasiveness and Clarity | Rozenblit and Keil 2002 | Explanation can be persuasive even if it does not increase understanding (illusion of explanatory depth). |

# Central concepts for human-centered XAI

Lessons for developing human-centered XAI have emerged across a wide range of literature, extending back to the 1970s. In Table 1, we identify some of the major concepts from this literature, ranging from aspects of measurement (how do you assess explanation), to components of system architectures (what must it represent and what interactions must it support), to forms of explanations that have been shown to be helpful. Each of these represents a principled assertion that has been made by one or more (often many) researchers across several subdomains, often with substantial empirical support.

## Measurement Domains

Hoffman et al. (2018a) proposed a descriptive model of explanation in AI that distinguishes several measurement approaches for evaluating the effectiveness of explanation. One grouping is a variety of measures referred to as *explanation goodness* (see also Mueller et al, 2019). These are measures for assessing an explanation that are either objective or can be examined independently of a user. For example, Swartout and Moore (1993) proposed that fidelity, understandability, and sufficiency were properties of good explanations. Similarly, Kulesza et al. (2015) identified several goodness criteria, including iterativity, soundness, completeness, non-overwhelmingness, and reversibility; Ribera and Lapedriza (2019) suggested that explanations should follow Grice's (1975) maxims; and Wang et al (2019) referred to this approach as 'unvalidated guidelines'. Hoffman et al. (2018a) provided an explanation goodness checklist that might serve as one means of evaluating this construct.

In contrast, *explanation satisfaction* measures must be assessed in the context of a user trying to understand a system or achieve a task. Satisfaction involves assessments by the user (including trust, reliance, and similar concepts). Hoffman et al. (2018a) provided validated scales that could be used for measuring satisfaction, although many methods are available, especially in the context of trust in automation (see Adams et al. 2003, for a review).

Explanation should not be expected to simply increase trust and other satisfaction measures—it should only do so in a context in which it is *justified*. Hoffman (2017) discussed this issue, distinguishing between justified and unjustified trust and mis-trust. Both justified trust and mis-trust are positive outcomes of a good explanation, because it allows a user to know which functions can be assigned to automation, and which cannot be. Notions of justified (sometimes called "calibrated" or "appropriate" trust) have been explored at least since Muir's (1994; Muir and Moray 1996) research; and more recently Fallon and Blaha (2018) suggested that "transparency" is an important precursor to appropriate trust in machine learning systems.

Explanation should also improve a human user's *understanding* or *mental model* (see Forrester 1961) of a system. Assessing this mental model usually requires specific knowledge elicitation tailored to the circumstance (see Rouse and Morris 1986; Doyle and Ford 1998; Klein and Hoffman 2008). To test mental models, Kulesza et al. (2015) used direct knowledge tests, whereas Mueller et al. (2020) had participants give reasons an AI system might fail. Improved knowledge is expected to enhance other factors, including trust, satisfaction, and system performance.

Another consequence of an effective explanation is i*mproved performance* by the human-AI system. In the context of studying human-autonomy trust, Muir (1994) proposed that system performance is a useful outcome of enhanced trust; similarly, Muir and Neville (1996) demonstrated that performance is distinct from trust, insofar as trust could be reduced by errors that do not impact overall performance. In addition, Karsenty and Brézillon (1995) suggested that the goal of explanation in human-machine systems is to improve cooperative problem solving—a type of performance measure. Consequently, performance is an important potential outcome of an explainable system.

Finally, we note that for prospective use of an AI system that might learn and change over time, Tate et al. (2016) have advocated a *licensure* framework—identifying capabilities for which a system is permitted to be used in the future. It is feasible that certain capabilities of automation can only pass licensure if they have sufficient explanatory capabilities. It is also noteworthy that many evaluations of XAI have specifically addressed competency testing as a mode of performance, which is related to licensure.

## Explanation Architecture Features

Although explanations have been proposed that take on many different forms, researchers have repeatedly discovered and advocated a common set of properties that an explanation 'architecture' might support.

Many of these features are based on the finding that good explanations depend on context. For example, Kass and Finin (1988) advocated for *user models*—a perspective going back at least to Conant and Ashby (1970). A user model might allow an AI system to tailor an explanation to a particular kind of user, or tailor the explanation to a long-term interaction in which the user is learning over time (see Naiseh et al. 2020). Similarly, Brézillon (1994) and Doyle et al. (2003) have advocated for other aspects of *context*, including understanding goals, the environment, the user's expertise, and so on. Although this appears obvious, it is remarkable that most proposed XAI systems ignore this and offer a one-size-fits-all explanation based on a data-centric algorithm. In this approach, the XAI generates a generic explanation, it is delivered to the user, and it is simply assumed it will make the system explainable.

Another aspect of context is that explanation systems need to provide explanations *relevant* to the user's goals and knowledge. Explanations are often not necessary, but become useful in response to specific triggers, such as surprise or a violation of expectation (Lim and Dey 2009; Hoffman et al. 2018a). Systems that can anticipate the appropriate triggers for explanation will be more responsive to user needs. In discussing case-based XAI, Sørmo et al. (2005) pointed to Leake's (1995) work on abductive reasoning, which argued that "when to explain" and "what to explain" are central aspects of abductive reasoning (see also Hoffman et al. 2020, for an argument linking explanation to abduction). Similarly, explanations need to be *timely.* For example, Alam (2020) found that physicians used different explanations at different points in a diagnostic process, and found that explanations only impacted trust and satisfaction of users of a simulated AI system at critical points during the diagnosis—when errors were occurring or diagnoses were changing. An explanation architecture that is aware of this context will provide an overall more useful system. Timeliness and relevance may be difficult to anticipate by an algorithm, but this might be solved by building systems and architectures that have users drive and request explanations when they are needed.

In the context of teaching users about intelligent software, Mueller and Klein (2011) argued for using *experiential* methods—allowing users to experience a tool, observe success and failures, and especially learn about the boundary conditions of the tool. Similarly, Lakkaraju et al. (2017) argued that explanations should be *explorable, iterative*, and *interactive* as well, resulting in an explanation interface that allows users to answer their own questions.

## Properties and Formats of Explanations

The previous section identified basic properties of an AI explanation system, without addressing the form or content of explanations directly. In this paper, we do not focus on presenting a taxonomy of algorithms or approaches that have been tried or implemented to support explanation (as done by Chari et al. 2020). Rather, we focus on higher-level psychological considerations that can serve as general templates for effective explanations. Most algorithmic approaches fit one or more of these categories.

One such category, which has been referred to as *focus*, is whether the explanation is about how the system works (global) versus why a decision was made (*local*, but sometimes described as a *justification*). Both appear important, and although the European Union's "right to explanation" (Goodman and Flaxman 2017) focuses on local ("understand why a particular decision was reached"), other policies may favor global accounts (those that advocate for understanding how a system works). Recent research has appeared to focus on local justifications, although one famous exception is Zeiler et al.'s (2014) visualization of internal deep learning network layers. It is also true that global understanding of any complex system can, and often does, come from a mixture of global and local explanatory material. Little research has contrasted the relative benefits of global and local explanation, but Alam (2020) showed that local justification improved immediate satisfaction of decisions, whereas global explanation impacted final judgments of understanding.

In early XAI systems, explanations tended to take on the form of *logical rationales* (see Swartout 1977). These still appear in modern systems (see Hendricks et al. 2016), often in conjunction with other methods. Alam (2020) found that rationales were an important mode of explanation used by physicians in diagnosis, but also found that written rationales did not improve user's satisfaction in a simulated diagnosis system, suggesting that other formats such as visualizations and examples may be more powerful.

A more common approach in recent systems relies on *examples and cases* (see Doyle et al. 2003). One thing examples can do is to help establish a pattern, which can be important for a user who is first experiencing a system. However, it is not obvious how examples should be chosen. Yang and Shafto (2017) identified a Bayesian approach for selecting optimal examples to help teach categories, and Kim et al. (2016) used an example-based approach that highlighted both typical and atypical examples of a class. Examples are a starting point for explanation, but the examples must be chosen so that they help a user understand a category/concept or a decision boundary; randomly-chosen examples may not be helpful at all.

The fact that example-based approaches often use comparison is a specific instance of a general principle of *contrast and counterfactual reasoning* in XAI (Miller 2017). Contrast need not be based only on examples, and can take many forms. In general, a contrast is simply an implementation of a scientific experiment: if only one thing is varied and produces a difference, we can conclude that it caused the difference. The dimensions that are contrasted vary considerably in XAI, from examples (positive versus negative), to features (presence or absence), to saliency approaches (identifying whether a feature or region of an image was used in a decision), to counterfactuals (what a system would do if a particular feature were changed).

In fact, many *feature-highlighting approaches* (e.g., LIME: Ribeiro et al. 2016; GRADCAM: Selvaraju et al. 2017) are contrast approaches applied to a feature space. These approaches use heat maps (sometimes called "saliency maps") to highlight regions of a feature space that were critical for a decision. This can be done by an occlusion-based method: a critical region can be identified by determining how the decision changes when different regions of an image are erased or occluded. If a decision changes substantially, this suggests the occluded region is critical for the decision. These systems are sometimes considered localization algorithms, but highlighting of spatial features is essentially localization.

Another kind of contrast relates to the ***boundary conditions*** of the algorithm. Any AI system will have regions of competence and incompetence (Mueller et al. 2020), and we have suggested that experiential training can be helpful at understanding the competency envelope of a system, by showing both how the system works and *how it does not work* (Mueller and Klein 2010). This approach maps onto global explanations of a system, but its experiential nature relies on cases and local explanations of situations to support this in a tutorial format.

Both AI and human education practice have been influenced by notions of "active learning". Active learning (see Settles 2009 in machine learning and Meyers and Jones 1993 in education) has had a much larger influence in education, but in both fields, it represents a shift in focus from one of the information a learner is expected to master, toward the learner and the activities the learner does to select or engage in the material being learned. Similarly, both AI and psychology have undergone active vision revolutions (see Aloimonos et al. 1988 in machine vision and Findlay, Findlay, and Gilchrist 2003 in human vision). These approaches consider the active role of the human, and we believe that the field of XAI can also benefit from a similar shift, toward a focus on ***active and self-explanation***.

In support of this, research from the educational literature has shown how explaining things to oneself or to another can be an important means of learning (see Chi et al. 1989), and researchers have considered explanation as a cooperative process between the human and the system (Brézillon 1994). Indeed, Kaur et al (2020) discovered that users typically did not use interpretability tools as intended by their designers, suggesting their need for self-explanation is paramount. Related to this is the role of critique in developing an understanding of a system. Active explanation can be guided by critique processes, and research has shown that critiqueing can be an effective way of improving planning and forecasting (Veinott, Klein, and Wiggins 2010). Automated critic-based approaches have been proposed for XAI systems (see Langlot and Shortliffe 1989; Guerlain 1995). In such approaches, a plan or decision is analyzed and critiques are elicited that, for example, compare risks of the suggested plan with alternatives. Here, the XAI relies on the reasoning of the user in a sort of Socratic dialog to encourage self-explaining.

Finally, we note that explanation is linked closely to many areas of research on ***persuasion***, graph literacy, and information visualization (i.e., ***clarity***). This research is relevant in that it shows the impact of information (the explanation) depends on format, social factors, and individual differences in education and experience of the explainee. This needs to be leveraged in order for explanations to be effective, but this also suggests that the same factors that enhance persuasiveness may be mis-used: convincing a user that a system is trustworthy even if that is unjustified. One example of this is the so-called "illusion of explanatory depth" (Rozenblit and Keil 2002), which describes how people can be overconfident about their understanding of complex systems. Providing more explanatory material may simply lead people to feel that the system is more trustworthy, and might not actually lead to an increase in understanding or an improvement in performance.

Next, we present two practical outcomes of the concepts reviewed in Table 1. First, we describe the "Self-explanation Scorecard"--a structured means of evaluating a system to help identify strengths and weaknesses with respect to the ability of the system to support self-explanation. Then, we will identify a set of general design principles that may be useful for XAI engineers to guide their development of human-centered explanation systems for AI.

## The Self-explanation Scorecard

Klein et al. (2019) described a "Self-explanation scorecard", which is a set of capabilities provided by an AI systems that specifically support self-explanation.

The Scorecard is an ordinal ranking of explanation types that map generally on to the complexity of the required reasoning that it supports. This order is not an assertion about which levels are likely to be most effective for any specific case, or that particular explanation at a higher level will necessarily be more effective than another explanation at a lower level. However, we do suppose that in general, higher-level systems provide more relevant information to support the user in their attempt to self-explain. The levels of self-explanation and examples are described in Table 2.

Although many AI systems have no explicit support for explanation (***null***), many expose different ***features*** to help understand the analyses done by the AI. A user will typically self-explain by blending these features with other information at higher levels of the scorecard. An AI system that shows ***examples of successes*** supports self-explanation because it grounds abstract processing in specific cases. AI systems that expose ***decision logic*** can help the user understand rules for why an outcome occurred. ***Failures*** show places where the AI does not work, and can help understand limits of the system. Any of these simpler levels can form the basis for ***comparisons***, contrasts that help a user make sense of critical factors in the AI. Finally, explanations that help users ***diagnose errors and failures*** might be the most powerful, as the enable understanding of the contrasts that make the difference between success and failure.

Table 2. Levels of the self-explanation scorecard.

| 1. Null | No explanation (typical AI) |
|---|---|
| **2. Features** | Heat maps, bounding boxes, linguistic features, semantic bubbles illustrate some of the analyses done by the AI. Surface features by themselves don't help much in understanding how the AI works, but in conjunction with positive cases and failures they can be useful. |
| **3. Successes** | Instances or demonstrations of the AI generating decisions, responses, or recommendations. |
| **4. Mechanisms** | Global descriptions of how the AI works can refer to mechanisms or architecture. Typically is text, but may include example instances. This form of explanatory information is typically included in the initial instructions about the XAI system and its uses. |
| **5. AI Reasoning** | These are ways to "look under the hood" of the AI to get some idea of how it is making decisions. This can be shown via choice logic, decision rules, goal stacks, parse graphs. These can show the ways the AI weighs different pieces of information in order to make a choice. |
| **6. Failures** | Instances of AI mistakes breakdowns. These are often very informative as they illustrate limitations of the AI and also illustrate how the AI works (and doesn't work). Failures can also be with respect to the explanations, i.e., user feedback to the AI about whether an explanation is correct. |
| **7. Comparisons** | Comparisons can be expressed using analogs (highlighting similarities and differences.) or counterfactuals. Comparisons can contrast choice logic or factor weights for different conditions or for successes vs failures. |
| **8. Diagnoses of failures** | These are even more informative than the failures alone, they are Description of the reasons for failures. In addition, letting the user manipulate the AI and to infer diagnoses; use of manipulations allows users to create failure conditions and to make their own inferences about diagnoses. |

We have found this Scorecard useful for formative evaluation of XAI systems, in order to help developers clarify and understand their approach, their strengths, and their potential weaknesses.

## Design Principles for Human-centered Explainable Systems

Table 1 distilled a range of scholarship that has established architectures, properties, and measurement approaches that have been used to develop explainable systems in the past. However, simply applying these patterns to new systems is no guarantee for success, and not all of the principles, as stated, are immediately actionable. These are measures, architecture features, and explanation types that have been shown to be effective in some contexts, but they are likely to fail in others. Any strong conclusion about explainability must be based on empirical evidence, and requires testing in the context of actual work with the intended representative users of the system.

Nevertheless, we may be able to identify some of the most central principles that should be observed when developing human-centered XAI. These systems may indeed start with an algorithmic explanation, but in order to succeed, will need to work in the context of a work domain to help the user achieve their goals. Some of the principles we have identified include:

- **The property of being an explanation is not a property of statements, visualizations, or examples.** Explaining is a process by which an explainee and explainer achieve common ground. XAI Designers should recognize that an explanation is not merely an artifact delivered from an algorithm, but must be understood by a user to be effective.
- **Work matters.** It is impractical to develop a useful and usable explanation system outside of a work context. Explanation relates the tool to the user knowledge in the context of the goals and tasks, and whether a particular algorithm or visualization will support that work cannot be known in the abstract. Human-centered design principles suggest involving the user early and throughout the system development process (Greenbaum and Kyng 1991; Deal and Hoffman 2010; Hoffman et al. 2010; Clancey 2020).
- **The importance of active self-explanation.** The "spoon feeding" paradigm is oblivious to the fact that irrespective of whatever material people receive by way of explanation, they still engage in a motivated attempt to make sense of the AI and the explanatory material. Developers should recognize this and focus explanatory systems on information that empowers users to self-explain, rather than simply delivering an output of an algorithms that is intended to serve as explanatory.
- **Build explanatory systems, not explanations.** Rarely does the initial explanation coming directly from the AI algorithms provide a useful explanation, let alone an ideal one. It must be accompanied by other things (instructions, tutorial activities, comparisons, exploratory interfaces, user models, etc.) to succeed.
- **Combined methods are necessary.** Much work on XAI involves testing a particular concept or algorithm in isolation. But when designing an explanatory system, multiple kinds of information can complement one another. For example, both global and local explanations may be justifiable and reinforce one another. Showing examples that establish a pattern may play a different role than contrasting examples which establish a critical causal relationship. Using examples of related heatmaps may be more powerful than either examples or heatmaps in isolation. And it is crucial to keep in mind that actual work contexts involve multiple systems, so the user is continuously challenged with their own "system integration" task.
- **An explanation can have many different consequences.** Often, developers create and test explanations to determine whether they work. However, different explanations can have very different effects. This includes differing effects on qualitative assessments (satisfaction, trust), versus knowledge measures and performance measures. The explanations should be tuned to the goal, keeping in mind the fact that people may be led to trust and rely upon an AI system simply being given more and more information about it, whether or not that information leads to better or deeper understanding.
- **Measurement Matters.** Because explanations can have their impact on a number of ways, and so can be assessed along many dimensions (goodness, satisfaction and trust, knowledge/understanding, and performance). Designers should identify what consequences the explanation should have in order to develop an appropriate measurement and assessment approach.
- **Knowledge and understanding are central.** Much of the research on XAI focuses on algorithmic visualizations, which distracts from the fact that the focus of explanation is on developing a better understanding of the system. Understanding leads to appropriate trust and appropriate reliance, and therefore overall better performance with the system.
- **Context matters: Users, timing, goals.** An explanation is not a beacon revealing the truth. The best explanation depends on context: who the user is, what their goal is, when they need an explanation, and how its effectiveness is measured. Developers should consider use cases, user models, timeliness, and attention and distraction limitations for their explanations.
- **The power of differences and contrast.** A central lesson of XAI is the utility of contrast, comparison, and counterfactuals in understanding the boundary conditions of a system. A useful exercise is to first develop learning objectives for an explanatory system, and identify the contrasts necessary to support those objectives.
- **Explanation is not just about transparency.** If something is transparent, you cannot see it. This word is widely misused. What is needed are systems whose workings are *apparent*, that is, readily understood and not hidden (the "black box metaphor"). Especially in the context of fairness in algorithmic decision making, many have advocated "transparency" as an approach to explanation. This can never be enough, because a user may still not understand how a system works even if its algorithms are somehow apparent---observable and visible. Other methods (contrast, global explanation and local justification, examples, explorable interfaces that permit hypotheticals, etc.) will be necessary in most situations to harness apparency and develop understanding. In "real world" work contexts, people always feel some mixture of justified trust, unjustified trust and justified and unjustified mistrust. These attitudes are in constant flux and rarely develop in a smooth progression toward some ideal and stable point. Trust can come and go in a flash. When the AI fails in a way that a human would never fail, reliance can collapse.
- **The need for explanation is "triggered".** Too often, XAI systems deliver an explanation regardless of whether one is needed. However, explanations are not always necessary. In normal human reasoning, explanation is triggered by states such as surprise and violations of expectation. Advances in XAI will come when systems begin to understand situations that are likely to engender surprise and violate user expectations.
- **Explanation is knowledge transformation and sense-making.** The achievement of an understanding is not just the learning or incorporation of information; it involves changing previous beliefs and preconceptions. The acquisition of knowledge involves both assimilation and accommodation, to use Piagetian terminology. The power of explanation is that it can activate fast "System 2" learning modes that quickly reconfigure knowledge with minimal feedback, bypassing the slower "System 1" trial-and-error feedback-based learning often used to understand a system. XAI systems should harness this by attempting to identify the user's current understanding (so that it can better predict how to transform this knowledge), and support the information that will help make these transformations.

- **Explanation is never a "one-off."** Especially for AI systems that learn or are applied in dynamic contexts, users often need repeated explanations and re-explanations. How has the algorithm changed? Are these new data valid? XAI systems might benefit from considering the long-term interaction with users, even in simple ways like recognizing that once learned, an explanation may not need to be given again unless something has changed.

## Summary

In this paper, we have identified a broad set of concepts and principles that have arisen in search of better user-centered XAI systems, including measurement approaches, elements of explainable architectures, and formats of explanation. These represent the findings of a broad range of research and scholarship at the nexus of AI, automation, human-systems integration, cognitive systems engineering, and psychology. Our main conclusion is that XAI must be understood as a problem not just for AI developers, but from the perspectives of human cognition, and the emergent human-machine cognitive systems. The Self-explanation Scorecard and list of design principles for human-centered XAI are two tools that XAI developers can use to guide development of useful, understandable, and usable AI systems.

## Acknowledgements

This research was developed with funding from the Defense Advanced Research Projects Agency (DARPA). The views, opinions and/or findings expressed are those of the author and should not be interpreted as representing the official views or policies of the Department of Defense or the U.S. Government.